%% file: main.tex
\documentclass{article}

\usepackage[preprint]{nips_20  18}

\usepackage[utf8]{inputenc} % allow utf-8 input
\usepackage[T1]{fontenc}    % use 8-bit T1 fonts
\usepackage[english]{babel}
\usepackage{hyperref}       % hyperlinks
\usepackage{url}            % simple URL typesetting
\usepackage{booktabs}       % professional-quality tables
\usepackage{amsmath}
\usepackage{amsthm}
\usepackage{amssymb}
\usepackage{amsfonts}       % blackboard math symbols
\usepackage{nicefrac}       % compact symbols for 1/2, etc.
\usepackage{microtype}      % microtypography
\usepackage[colorinlistoftodos]{todonotes}
\usepackage{graphicx}
\usepackage{natbib}
\usepackage{relsize}
\usepackage{xfrac}
\usepackage{tikz}
\usepackage{subcaption} 
\usepackage{cleveref}
\usepackage[]{algorithm2e}
\usepackage[title]{appendix}
\usepackage{subcaption}

\title{Taming VAEs} 

\usepackage{macros}

\author{
  {Danilo J. Rezende}
  \thanks{Both authors contributed equally to this work.}
  \And
  {Fabio Viola}
  \footnotemark[1]
  \AND
  {\tt \{danilor, fviola\}@google.com} \\
  DeepMind, London, UK
}

\begin{document}

\maketitle

\input{abstract.tex}

\input{introduction.tex}

\input{related_work.tex}

\input{methods.tex}

\input{experiments.tex}

\input{discussion.tex}

\bibliographystyle{unsrt}
\bibliography{references}

\clearpage
\begin{appendices}
\crefalias{section}{appsec}
\crefalias{subsection}{appsec}
\input{appendix.tex}
\end{appendices}

\end{document}

%% file: abstract.tex
\begin{abstract}
In spite of remarkable progress in deep latent variable generative modeling, training still remains a challenge due to a combination of optimization and generalization issues. 
In practice, a combination of heuristic algorithms (such as hand-crafted annealing of KL-terms) is often used in order to achieve the desired results, but such solutions are not robust to changes in model architecture or dataset. The best settings can often vary dramatically from one problem to another, which requires doing expensive parameter sweeps for each new case.
Here we develop on the idea of training VAEs with additional constraints as a way to control their behaviour. 
We first present a detailed theoretical analysis of constrained VAEs, expanding our understanding of how these models work. We then introduce and analyze a practical algorithm termed {\it \badaptlong{}}, \badaptshort{}. The main advantage of \badaptshort{} for the machine learning practitioner is a more intuitive, yet principled, process of tuning the loss. This involves defining of a set of constraints, which typically have an explicit relation to the desired model performance, in contrast to tweaking abstract hyper-parameters which implicitly affect the model behavior. Encouraging experimental results in several standard datasets indicate that \badaptshort{} is a very robust and effective tool to balance reconstruction and compression constraints.
\end{abstract}

%% file: introduction.tex
\section{Introduction}\label{sec.intro}

Deep variational auto-encoders (VAEs) have made substantial progress in the last few
years on modelling complex data 
% \cite{Kingma2013AutoEncodingVB,Rezende2014StochasticBA,Kingma2016ImprovedVI,Snderby2016LadderVA,Gregor2016TowardsCC,Gulrajani2016PixelVAEAL,Bachman2016AnAF, akuzawa2018expressive,hsu2017learning, gomez2016automatic, sultan2018transferable,hernandez2017variational,inoue2017transfer, park2018multimodal,van2016stable,GQN,rezende2016unsupervised,eslami2016attend}.
such as images \cite{Kingma2013AutoEncodingVB,Rezende2014StochasticBA,Kingma2016ImprovedVI,Snderby2016LadderVA,Gregor2016TowardsCC,Gulrajani2016PixelVAEAL,Bachman2016AnAF}, speech synthesis \cite{akuzawa2018expressive, hsu2017learning}, 
molecular discovery \cite{gomez2016automatic, sultan2018transferable, hernandez2017variational}, 
robotics \cite{inoue2017transfer, park2018multimodal, van2016stable} and
3d-scene understanding \cite{GQN, rezende2016unsupervised, eslami2016attend}.

VAEs are latent-variable generative models that define a joint density $p(\vx, \vz)$ between some observed data $\vx \in \RR^{d_x}$ and unobserved or latent variables $\vz \in \RR^{d_z}$. The most popular method for training these models is through stochastic amortized variational approximations \cite{Rezende2014StochasticBA, Kingma2013AutoEncodingVB}, which use a variational posterior (also referred to as encoder), $q(\vz| \vx)$, to construct the evidence lower-bound (ELBO) objective function. 

One prominent limitation of vanilla VAEs is the use of simple diagonal Gaussian posterior approximations. It has been observed empirically that VAEs with simple posterior models have a tendency to ignore some of the latent-variables (latent-collapse) \cite{Tomczak2018VAEWA,Burda2015ImportanceWA} and produce blurred reconstructions \cite{Burda2015ImportanceWA, Snderby2016LadderVA}. As a result, several mechanisms have been proposed to increase the expressiveness of the variational posterior density \cite{Nalisnick2016ApproximateIF, Rezende2015VariationalIW,Salimans2015MarkovCM,Tomczak2016ImprovingVA,Tran2015TheVG, Kingma2016ImprovedVI,Berg2018SylvesterNF,Gregor2016TowardsCC} but it still remains a challenge to train complex encoders due to a combination of optimization and generalization issues \cite{Cremer2018InferenceSI,Rosca2018DistributionMI}. 
Some of these issues have been partially addressed in the literature through heuristics, such as hand-crafted annealing of the KL-term \cite{GQN,Snderby2016LadderVA,Gregor2016TowardsCC}, injection of uniform noise to the pixels \cite{theis2015note} and reduction of the bit-depth of the data. 
A %particular 
case of interest are VAEs with information bottleneck constraints such as $\beta$-VAEs \cite{Higgins2016vaeLB}. While a body of work on information bottleneck has primarily focused on tools to analyze models \cite{Alemi2017FixingAB,tishby2000information,tishby2015deep}, it has also been shown that VAEs with various information bottleneck constraints can trade off reconstruction accuracy for better-factorized latent representations \cite{Higgins2016vaeLB,Burgess2018UnderstandingDI,Alemi2016DeepVI}, a highly desired property in many real-world applications as well as model analysis. Other types of constraints have also been used to improve sample quality and reduce latent-collapse \cite{zhao2017towards}, but while using supplementary constraints in VAEs has produced encouraging empirical results, we find that there have been fewer efforts towards general tools and theoretical analysis for constrained VAEs at scale. 

Here, we introduce a practical mechanism for controlling the balance between compression (KL minimization) and other constraints we wish to enforce in our model (not limited to, but including reconstruction error) termed {\it \badaptlong{}}, \badaptshort{}. \badaptshort{} enables an intuitive, yet principled, work-flow for tuning loss functions. This involves the definition of a set of constraints, which typically have an explicit relation to the desired model performance, in constrast to tweaking abstract information-theoretic hyper-parameters which implicitly affect the model behavior. In spite of its simplicity, our experiments support the view that \badaptshort{} is an empowering tool and we argue that it has enabled us to have an unprecedented level of control over the properties and robustness of complex models such as ConvDraw \cite{Gregor2016TowardsCC, GQN} and VAEs with NVP posteriors \cite{Rosca2018DistributionMI}.

The key contributions of this paper are: 
(a) With a focus on ELBOs with supplementary constraints, we present a detailed theoretical analysis of the behaviour of high-capacity VAEs with and without KL and Lipschitz constraints, advancing our understanding of VAEs on multiple fronts: (i) We demonstrate that the posterior density of unconstrained VAEs will converge to an equiprobable partition of the latent-space; (ii) We provide a connection between $\beta$-VAEs and spectral clustering; (iii) Drawing from statistical mechanics we study phase-transitions in the reconstruction fixed-points of $\beta$-VAEs; (b) Equipped with a better understanding of the impact of constraints in VAEs, we design the \badaptshort{} algorithm.
% \end{itemize}

The remainder of the paper is structured as follows: in \Cref{sec.tiles} we study the behavior of unconstrained VAEs at convergence; in \Cref{sec.kernels} we draw links between $\beta$-VAEs and spectral clustering, and we study phase-transitions in $\beta$-VAEs; in \Cref{sec.agbo} we present the \badaptshort{} algorithm; finally, in \Cref{sec.experiments} we illustrate our theoretical analysis on mixture data and we assess the behavior of \badaptshort{} on several larger datasets and for different types of constraints.

%% file: related_work.tex
\section{Related work} \label{sec.relatedwork}
,Similarly to \cite{zhao2017towards} we study VAEs with supplementary constraints in addition to the ELBO objective function and we study its behaviour in the information plane as in \cite{Alemi2017FixingAB}. Our theoretical analysis extends the analysis performed in \cite{zhao2017towards, Alemi2017FixingAB} with an emphasis on the properties of the learned posterior densities in high-capacity VAEs.
Our analysis also extends the idea of information bottleneck and geometrical clustering from \cite{strouse2017information} to the case where latent variables are continuous instead of discrete.
A further extension of the analysis including the incorporation of Lipschitz constraints is also available in \Cref{sec.lipschitz}. 
% and through the incorporation of Lipschitz constraints.
Complementary to the analysis from \cite{Dai2017HiddenTO} with (semi-)affine assumptions on the VAE's decoder, we focus on non-linear aspects of high-capacity constrained VAEs.

% To obtain a tractable representation of both the posterior density and decoder for propositions \eqref{prop.tiling} and \eqref{prop.kernel}, we express them in a particular functional basis that is analogous to using a mixture posterior density as in \cite{Nalisnick2016ApproximateIF} but where the modes of the mixture have non-overlapping supports. This assumption substantially simplify our analysis, allowing us to reach meaningful conclusions.

\badaptshort{} is a simple mechanism for approximately optimizing VAEs under different types of constraint. It is inspired by the empirical observation that some high-capacity VAEs such as ConvDRAW\cite{Gregor2016TowardsCC, GQN} may reach much lower reconstruction errors compared to a level that is perceptually distinguishable, at the expense of a weak compression rate (large KL). \badaptshort{} is designed to be an easy-to-implement and tune approximation of more complex stochastic constrained optimization techniques \cite{wang2003stochastic, rocha2010stochastic} and to take advantage of the reparametrization trick in VAEs \cite{Kingma2013AutoEncodingVB,Rezende2014StochasticBA}. At a high-level, \badaptshort{} is similar to information constraints studied in \cite{Alemi2017FixingAB, phuong2018the, Zhang2017InformationPA, Kolchinsky2017NonlinearIB}. However, we argue that in many practical cases it is much easier to decide on useful constraints in the data-domain, such as a desired reconstruction accuracy, rather than information constraints. Additionally, the types of information constraints we can impose on VAEs are restricted to a few combinations of KL-divergences (\eg mutual information between latents and data), whereas there are many easily available ways of meaningfully constraining reconstructions (\eg bounding reconstruction errors, bounding the ability of a classifier to correctly classify reconstructions or bounding reconstruction errors in some feature space).

Some widespread practices for modelling images such as injecting uniform noise to the pixels and reducing the bit-depth of the color channels (e.g. \cite{theis2015note, kingma2018glow, Gregor2016TowardsCC, Rosca2018DistributionMI}) can also be mathematically interpreted as constraints which bound the values of the likelihood from above. For instance, training a model with density $p(\vx)$ by injecting uniform noise to the samples $x \rightarrow x + b \epsilon, \epsilon \sim \text{unif}(-1/2, 1/2)$ is a way of maximizing the likelihood under the constraint $p(\vx) \leq \frac{1}{b}$.
With \badaptshort{}, there is no need to resort to these heuristics.

%% file: methods.tex
\section{Methods}\label{sec.methods}

VAEs are smooth parametric latent variable models of the form $p(\vx, \vz) = p(\vx | \vz) \pi(\vz)$ and are trained typically by maximizing the ELBO variational objective, $\mathcal{F}$, using a parametric variational posterior $q(\vz | \vx)$.
\begin{align}
\mathcal{F} &= \EE{\rho(\vx)}{ \EE{q (\vz |  \vx)}{\ln p(\vx | \vz)} - \KL{q (\vz |  \vx)}{\pi (\vz)}}, \label{eq.elbo}
\end{align}
where the prior is taken to be a unit Gaussian $\pi(\vz)=\mathcal{N}(\vec{0}, \mathbb{I})$ and the empirical data density is represented as $\rho(\vx) = \frac{1}{n}\sum_i^n \delta(\vx - \vx_i)$, where $\vx_i$ are the training data-points.

We focus on VAEs with Gaussian decoder density of the form $p(\vx|\vz)=\mathcal{N}(\vx | g(\vz), \sigma^2 \mathbb{I})$, where $\sigma$ is a global parameter and $g(\vz)$ is referred to as a decoder or generator. Restricting to decoder densities where the components $x_i$ of $\vx$ are conditionally independent given a latent vector $\vz$ eliminates a family of solutions in the infinite-capacity limit where the decoder density $p(\vx|\vz)$ ignores the latent variables, i.e. $p(\vx|\vz)=p(\vx) \approx \rho(\vx)$, as observed in \cite{Alemi2017FixingAB}. This restriction is important because we are interested in the behaviour of the latent variables in this paper.

In contrast to ELBO maximization, we consider a constrained optimization problem for variational auto-encoders where we seek to minimize the KL-divergence, $\KL{q (\vz |  \vx)}{\pi (\vz)}$, under a set of expectation inequality constraints of the form $\EE{\rho(\vx) q (\vz |  \vx)}{\mathcal{C}(\vx, g(\vz))} \leq \vec{0}$ where $\mathcal{C}(\vx,g(\vz)) \in \RR^L$. We refer to this type of constraint as {\it reconstruction constraint} since they are based on some comparison between a data-point $\vx$ and its reconstruction.
We can solve this problem by using a standard method of Lagrange multipliers, where we introduce the Lagrange multipliers $\vb \in \RR^L$ and optimize the Lagrangian $\mathcal{L}_{\vb}$ via a min-max optimization scheme \cite{bertsekas1996},
\begin{align}
\mathcal{L}_{\vb} &= \EE{\rho(\vx)}{\KL{q (\vz |  \vx)}{\pi (\vz)}} + \vb^T\EE{\rho(\vx) q (\vz |  \vx)}{\mathcal{C}( \vx, g(\vz))}.\label{eq.lagrangian}
\end{align}
%Since we consider $\sigma$ to be a fixed parameter in most of this paper, we have omitted constant terms ($d_x \ln \sigma$) from the Gaussian likelihood.
A general algorithm for finding the extreme points of the ELBO $\mathcal{F}$ and of the Lagrangian $\mathcal{L}_{\vb}$ is provided in \Cref{prop.fixedpoint.eqs}, where we extend the well-known derivations from rate-distortion theory \cite{blahut1972computation,csisz1984information,cover2012elements,tishby2000information} to the case with reconstructions constraints.

\begin{proposition} \label{prop.fixedpoint.eqs}
(Fixed-point equations) The extrema of the ELBO $\mathcal{F}$ with respect to the decoder $g(\vz)$ and encoder $q(\vz | \vx)$ are solutions of the fixed-point \Cref{eq.fp.q.elbo,eq.fp.g.elbo} and for the Lagrangian $\mathcal{L}_{\vb}$, \Cref{eq.fp.q,eq.fp.g} respectively
\begin{align}
  q^t_{\mathcal{F}} (\vz |  \vx) &\propto \pi(\vz) e^{- \frac{\| \vx - g^{t-1}_{\mathcal{F}}(\vz)\|^2}{2\sigma^2} } \label{eq.fp.q.elbo}\\
  g^t_{\mathcal{F}}(\vz) &= \sum_i w_i^t(\vz) \vx_i \label{eq.fp.g.elbo}\\
  q^t_{\mathcal{L}_{\vb}} (\vz |  \vx) &\propto \pi(\vz) e^{-H(\vx, \vz)} \label{eq.fp.q}\\
  0 &= \sum_i q^t_{\mathcal{L}_{\vb}} (\vz |  \vx_i) \vb^TG(\vx_i, g^t_{\mathcal{L}_{\vb}}(\vz)) \label{eq.fp.g},
\end{align}
where $H(\vx, \vz) = \vb^T\mathcal{C}( \vx, g^{t-1}_{\mathcal{L}_{\vb}}(\vz))$, $w_i^t (\vz) = \frac{q^t_{\mathcal{F}} (\vz |  \vx_i)}{\sum_j q^t_{\mathcal{F}} (\vz | \vx_j)}$ and $G(\vx, g(\vz)) = \frac{\partial \mathcal{C}( \vx, g(\vz))}{\partial g(\vz)}$.
See proof in \Cref{proof.fixedpoint.eqs}.
\end{proposition}

In the following \Cref{sec.tiles,sec.kernels} we analyze properties of \Cref{eq.fp.q.elbo,eq.fp.g.elbo,eq.fp.q,eq.fp.g} constraints to gain a better understanding of the behaviour of VAEs. In \Cref{sec.lipschitz} we also provide additional analysis of the effect of local and global Lipschitz constraints for the interested reader.  

\subsection{Unconstrained VAEs}\label{sec.tiles}
In the unconstrained case, where we optimize the ELBO, we can derive a few interesting conclusions from \Cref{eq.fp.q.elbo,eq.fp.g.elbo}: (i) The global optimal decoder $g(\vz)$ is a convex linear combination of the training data of the form $g (\vz) = \sum_i w_i (\vz) \vx_i$; (ii) If we optimize the standard-deviation $\sigma$ jointly with the decoder and encoder, it will converge to zero; (iii) The posterior density $q (\vz |  \vx)$ will converge to a distribution with support corresponding to one element of a partition of the latent space. Moreover, the set of supports of the posterior density formed by each data-point constitutes a partition of the latent space that is equiprobable under the prior. These results are formalized in \Cref{prop.tiling}, where we demonstrate that a solution satisfying all these properties is a fixed point of \Cref{eq.fp.q.elbo,eq.fp.g.elbo}.

\begin{proposition} \label{prop.tiling}
  (High-capacity VAEs learn an equiprobable partition of the latent space): Let $q (\vz |  \vx_i) = \pi (\vz)
  \mathbb{I}_{\vz \in \Omega_i}/ \pi_i$, where $\pi_i=\EE{\pi}{\mathbb{I}_{\vz \in \Omega_i}}$ is a normalization constant, be the variational posterior density, evaluated at a training
  point $x_i$. This density is equal to the restriction of the prior $\pi (\vz)$ to a limited support $\Omega_i \subset \mathbb{R}^{d_z}$ in the latent space. $q(\vz | \vx_i)$ is a fixed-point of equations \eqref{eq.fp.q} for any set of
  volumes $\Omega_i$ that form  a partition of the latent space. Furthermore, the highest ELBO is achieved when the partition is equiprobable under the prior. That is, $\EE{\pi}{\mathbb{I}_{\vz \in \Omega_i}} = \EE{\pi}{\mathbb{I}_{\vz \in \Omega_j}}=1/n$, 
  $\mathbb{R}^{d_z} = \cup_i \Omega_i$ and
  $\Omega_i \cap \Omega_j = \varnothing$ if $i \neq j$. See proof in \Cref{proof.tiling}.
\end{proposition}

The fact that the standard deviation will converge to $0$ results in a numerically ill-posed problem as observed in \cite{Mattei2018LeveragingTE}. Nevertheless, the fixed-point equations still admits a stationary solution where the VAE becomes a mixture of Dirac-delta densities centered at the training data-points.

It is known empirically that low-capacity VAEs tend to produce blurred reconstructions and samples \cite{zhao2017towards, Higgins2016vaeLB}. Contrary to a popularly held belief (e.g. \cite{nowozin2016f}), this phenomenon is not caused by using a Gaussian likelihood alone: as observed in \cite{zhao2017towards}, this is primarily caused by a sub-optimal variational posterior. 
The fixed-point equation \eqref{eq.fp.g.elbo} provides a mathematical explanation for this phenomenon, generalizing the result from \cite{zhao2017towards}: the optimal decoder $g(\vz)$ for a given encoder $q (\vz |  \vx)$ is a convex linear combination of the training data. If the VAE's encoder cannot accurately distinguish between multiple training data-points, the resulting weights $w_i(\vz)$ in the VAE's decoder will be spread across the same data-points, resulting in a blurred reconstruction. This is formalized in \Cref{prop.blurredrecs}. 
\begin{proposition} \label{prop.blurredrecs}
(Blurred reconstructions) If the supports $\Omega_i$ from \Cref{prop.tiling} are overlapping (i.e. $\Omega_i \cap \Omega_j \neq \varnothing$ for $i \neq j$), then the optimal reconstruction at a latent point $\vz$ for a fixed encoder $q(\vz |  x_i) = \pi (\vz) \mathbb{I}_{z \in \Omega_i}/\pi_i$ will be the average of all data-points mapping to any of the overlapping basis weighted by the inverse prior probability of the respective basis.  See proof in \Cref{proof.blurredrecs}.
\end{proposition}

Another striking conclusion we can derive from \Cref{prop.fixedpoint.eqs,prop.blurredrecs} is that the support of the optimal decoder as a function of the latent vector will be concentrated in the support of the marginal posterior. In fact, if we revisit the proof of \Cref{prop.fixedpoint.eqs} in \Cref{proof.fixedpoint.eqs} when there are regions in the latent space where $q(\vz|\vx) \approx 0$ we notice that the decoder is completely unconstrained by the ELBO in these regions. We refer to this as the "holes problem" in VAEs. This problem is commonly encountered when using simple Gaussian posteriors (\eg \cite{Kingma2016ImprovedVI}).

\begin{figure}[!ht] 
\centering
\includegraphics[width=0.8\textwidth]{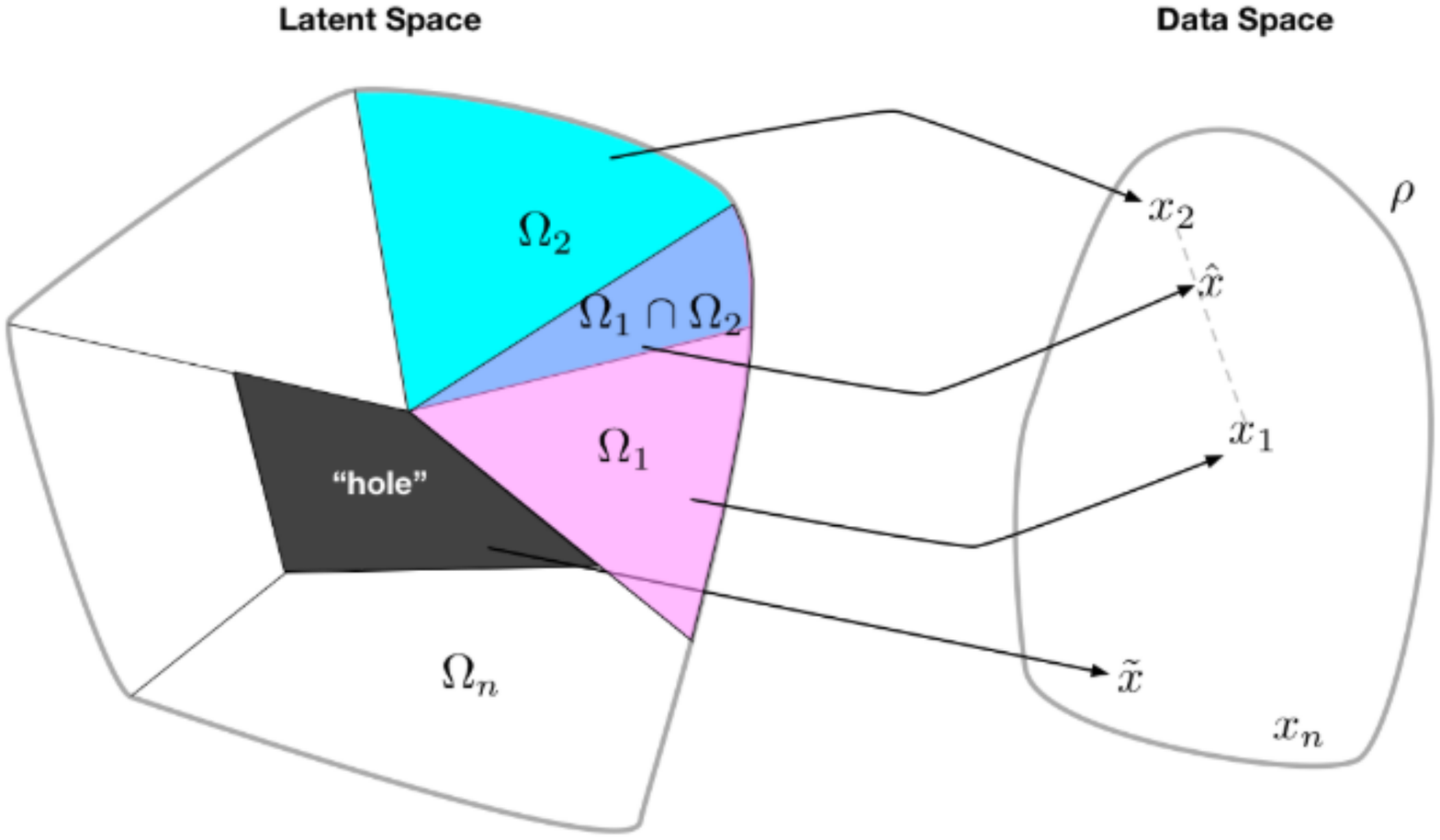}
\caption{
{\bf Illustration of the "blurred reconstructions" and the "holes" problems.} {\bf Left}: Latent space with a posterior with support in a tiling $\{\Omega_i\}$, where each tile $\Omega_i$ represents the support of the posterior for the data-point $x_i$.; {\bf Right}: Data space. In the region of the latent space where the posteriors of the data-points $x_1$ and $x_2$ overlap, $\Omega_1 \cap \Omega_2$, the optimal reconstruction $\hat{x}$ is a weighted average of the corresponding data-points, resulting in a blurred sample. In a region of low density under the marginal posterior, a "hole" (represented by the black area in the figure), the optimal reconstructions from these regions $\tilde{x}$ are unconstrained by the ELBO objective function.
}
\label{fig.blurred_holes}
\end{figure}

\subsection{High-capacity $\beta$-VAEs and spectral methods}\label{sec.kernels}

The $\beta$-coefficient in a $\beta$-VAE \cite{Higgins2016vaeLB} can be interpreted as the Lagrange multiplier of an inequality constraint imposing either a restriction on the value of the KL-term or a constraint on the reconstruction error \cite{Burgess2018UnderstandingDI, Alemi2017FixingAB}. When using the reconstruction error constraint $\mathcal{C}(\vx, g(\vz)) = \|\vx - g(\vz) \|^2 - \kappa^2$ in \Cref{eq.lagrangian}, the Lagrange multiplier $\vb$ is related to the $\beta$ from \cite{Higgins2016vaeLB} by $\vb = \frac{1}{ \beta}$.

While VAEs with simple linear decoders can be related to a form of robust PCA, \cite{Dai2017HiddenTO}, we demonstrate a relation between $\beta$-VAEs with high-capacity decoders and kernel methods such as spectral clustering. More precisely, we show in \Cref{prop.kernel} that the fixed-point equations of a high-capacity decoder expressed in a particular orthogonal basis are analogous to the reconstruction fixed point equations used in spectral clustering.

\begin{figure}[t] 
\centering
\includegraphics[width=\textwidth]{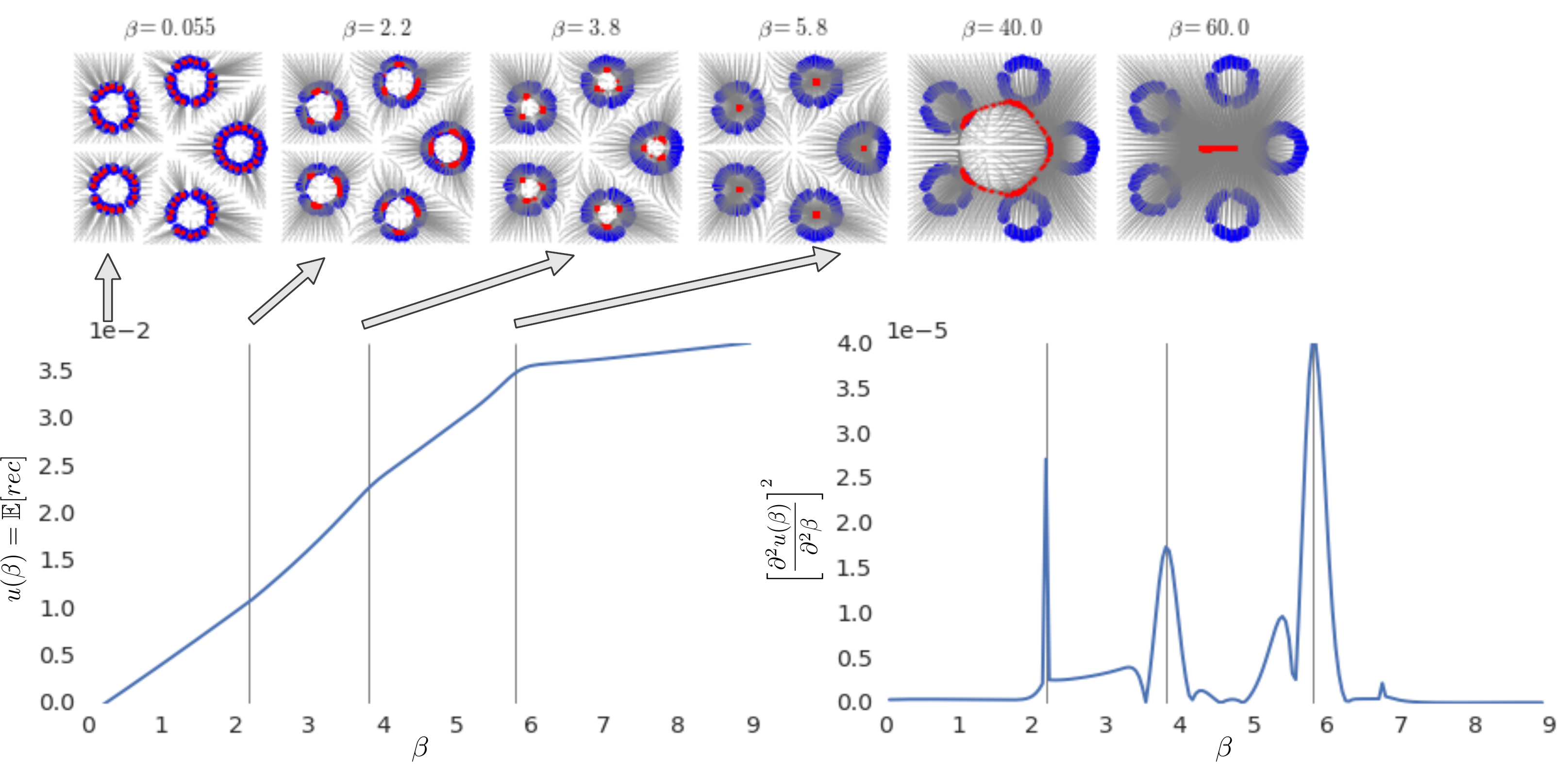}
\caption{
{\bf Effect of $\beta$ on the reconstruction fixed-points and phase-transitions}.  
{\bf Top images} Grey curves indicate the trajectories of the vectors $\psi$. Red points are the fixed-points. Blue points are the data points; {\bf Bottom left} Expected reconstruction error as a function of $\beta$. Vertical grey lines indicate the detected critical-temperatures $\beta_c^k$; {\bf Bottom right} Expected second-order derivative of the reconstruction error as a function of $\beta$. At critical temperatures reconstruction fixed-points will merge with each other, resulting in sudden changes in the slope of the reconstruction error with respect to the temperature $\beta$, these points correspond to spikes in the second-order derivatives. For analysis, we sorted the local maxima according to their height and restricted the analysis to the top-3 points, $\beta_c^{k=1,2,3}$. Details of this simulation are explained in \Cref{sec.experiments.fixedpoints}.
}
\label{fig.phasetransition}
\end{figure}

In the literature of spectral clustering and kernel PCA it is known that reconstructions based on a Gaussian Gram matrix may suffer phase-transitions (sudden change in eigen-values or reconstruction fixed-points) as a function of the scale parameter \cite{hoyle2004limiting, nadler2008diffusion,wang2012kernel,scholkopf1997kernel,mika1999kernel}. 
Making a bridge between VAEs and these methods allows us to investigate phase-transitions in high-capacity $\beta$-VAEs, where the expected reconstruction error is treated as an order parameter $u(\beta)=\EE{}{\|\vx - g(\vz) \|^2}$. In this case, phase transitions will occur at critical temperature points $\beta_c$, which can be detected by analyzing regions of high-curvature (high absolute second-order derivative $|\frac{\partial^2 u(\beta)}{\partial^2 \beta}|$). As in spectral clustering, $\beta$-VAEs phase transitions correspond to the merging of neighboring data-clusters at different spatial scales. This is illustrated in the experiment shown in \Cref{fig.phasetransition}, where we look at the merging of reconstruction fixed-points as we increase $\beta$. Interestingly, the phase-transitions that we observe are similar to what is known as first-order phase transitions in statistical mechanics \cite{blundell2009concepts}, but further analysis is necessary to clarify this connection.

\begin{proposition} \label{prop.kernel}
  (High-capacity $\beta$-VAE and spectral methods) Let $\phi_a : \mathbb{R}^{d_z}
  \rightarrow \{ 0, 1 \}$ be an orthogonal basis in the latent space. If we
  express the posterior density and generator using this basis respectively as $q (\vz | 
  x_i) = \pi(\vz) \sum_a m_{i a} \phi_a(\vz)$ and $g(\vz) = \psi^T \phi (\vz)$, where $m_{i a}$ is a matrix with
  positive entries satisfying the constraint $\sum_a m_{i a} \pi_a = 1$ where
  $\pi_a =\mathbb{E}_{\pi(\vz)} [\phi_a(\vz)]$. The fixed-point equations that
  maximize the ELBO with respect to $\psi$ are convergent under the appropriate initial conditions and are equivalent to computing
  the pre-images (reconstructions) of a Kernel-PCA model with a normalized Gaussian Kernel with scale parameter $\sqrt{\beta}$. See proof in \Cref{proof.kernel}.
\end{proposition}

\subsection{Equipartition of energy in $\beta$-VAEs}\label{subsec.equipartition}

In statistical mechanics there is an important result, known as {\it equipartition of energy theorem}. It states that, for a system in thermodynamic equilibrium, energy is shared equally amongst all accessible degrees of freedom at a fixed energy level, \cite{tolman1938principles}. We demonstrate that a similar theorem holds for VAEs in \Cref{prop.equipartition}. In \Cref{prop.tiling} we have seen that the optimal posterior for each data point in a high-capacity VAE has its support in the elements of a partition of the latent space and that this partition is equiprobable under the prior. We can generalize this result to $\beta$-VAEs by noticing that, as a result of the existence of reconstruction fixed points from \Cref{prop.kernel}, there will be regions in the latent space where the Hamiltonian $H(\vx, \vz)$ is approximately constant for a given $\vx$. At these regions, the posterior will be proportional to the prior and they will work as a discrete partition of the latent space, as in \Cref{prop.tiling}. The concept that VAE encoders learn a tiling of the latent space, each tile corresponding to a different level of the function $H(\vx, \vz)$, can be a guiding principle to evaluate generative models as well as to construct more meaningful constraints.
\begin{proposition}\label{prop.equipartition}
  (Equipartition of Energy for high-capacity VAEs). Let $H(\vx, \vz)$ be the "Hamiltonian" function from \Cref{prop.fixedpoint.eqs}
  for a VAE trained in a dataset with $n$ data-points. For a given data-point
  $\vx \in \RR^{d_x}$, latent point $\vz \in \RR^{d_z}$ and
  precision $\epsilon > 0$, let $\Omega (\vx, \vz_0) = \{ z' | | H (\vx,
  \vz') - H (\vx, \vz_0) | \leqslant \epsilon \} \subseteq \RR^{d_z}$. That
  is, $\Omega (\vx, \vz_0)$ is the set of latent points where the Hamiltonian is approximately constant. 
  As we vary $\vx$ and $\vz_0$, each set $\Omega (\vx, \vz_0)$ will be one element of a discrete set of disjoint sets, which we enumerate as $\Omega_a$. The encoder density $q (\vz | \vx_i)$ will converge to a mixture 
  of the restrictions of the prior to the basis elements
  $\Omega_{a}$. Moreover, the probability $\gamma_a$ of a sample from the prior falling in the partition element $\Omega_{a}$ is a solution of $\sum_i \frac{e^{- \frac{H_{i a}}{\beta}}}{\sum_b e^{- \frac{H_{i
b}}{\beta}} \gamma_b}=n$ where $H_{i a}$ is the value of $H(\vx_i, \vz)$ for $\vz \in \Omega_a$. See proof in \Cref{proof.equipartition}.
\end{proposition}

\subsection{The \badaptshort{} algorithm for VAEs}\label{sec.agbo}
To derive \badaptshort{} we start from the augmented Lagrangian defined in \Cref{eq.lagrangian} for a VAE with decoder parametrized by a vector $\theta$ and an encoder density parametrized by a vector $\eta$. Optimization of the loss involves joint minimization \wrt $\theta$ and $\eta$, and maximization \wrt to the Lagrange multipliers $\vb$. 
The parameters $\theta$ and $\eta$ are optimized by directly following the negative gradients of \Cref{eq.lagrangian}. The Lagrange multipliers $\vb$ are optimized following a moving average of the constraint vector $\mathcal{C}(\vx,g(\vz))$. 
In order to avoid backpropagation through the moving averages, we only apply the gradients to the last step of the moving average. This procedure is detailed in \Cref{alg.beta}. 

\begin{figure}[t] 
\centering
\includegraphics[width=\textwidth]{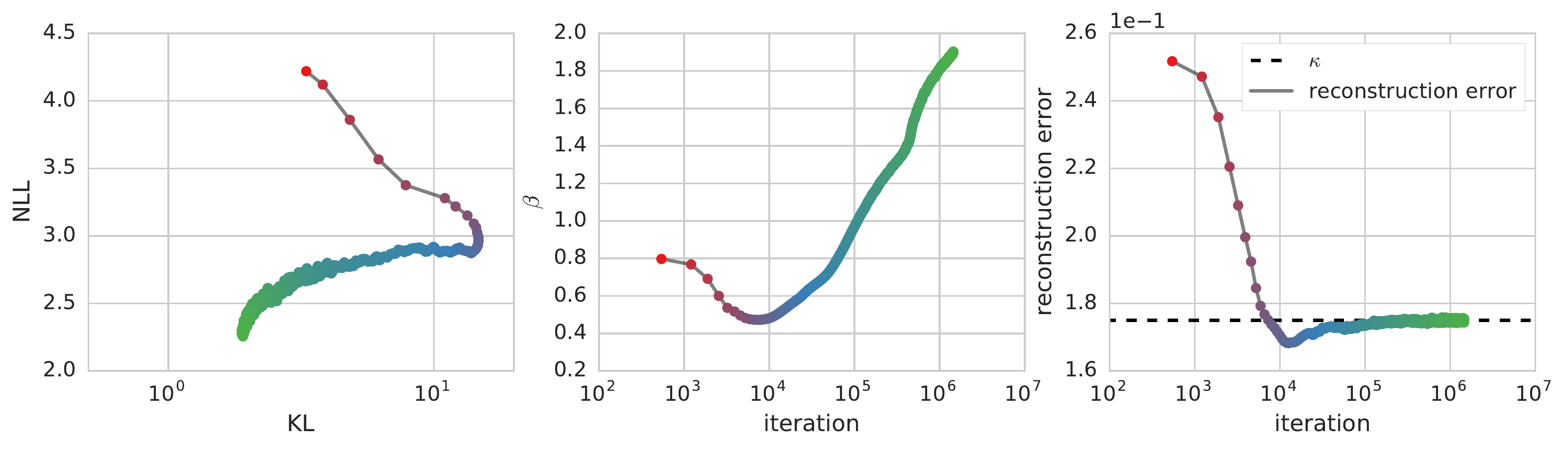}
\caption{
{\bf Trajectory in the information plane induced by \badaptshort{} during training}. This plot shows a typical trajectory in the NLL/KL plane for a model trained using \badaptshort{} with a RE constraint, alongside the corresponding values of the equivalent $\beta$ and pixel reconstruction errors; note that iteration information is consistently encoded using color in the three plots. At the beginning of training, $it < 10^4$, the reconstruction constraint dominates optimization, with $\beta < 1$ implicitly amplifying the NLL term in ELBO. When the inequality constraint is met, i.e. the reconstruction error curve crosses the $\kappa$ threshold horizontal line, $\beta$ slowly starts changing, modulated by the moving average, until at $it = 10^4$, the $\beta$ curve flexes and $\beta$ starts growing. This specific example is for a conditional ConvDraw model trained on MNIST-rotate.}
\label{fig.information_plane_trajectory}
\end{figure}

\Cref{fig.information_plane_trajectory} captures a representative example of the typical behavior of \badaptshort{}: early on in the optimization the solver quickly moves the model parameters into a regime of valid solutions, \ie parameter configurations satisfying the constraints, and then minimizes ELBO while preserving the validity of the current solution. We refer the reader to \Cref{fig.information_plane_trajectory}'s caption for a more detailed description of the optimization phases.

The main advantage of \badaptshort{} for the machine learning practitioner is that the process of tuning the loss involves the definition of a set of constraints, which typically have a direct relation to the desired model performance, and can be set in the model output space. This is clearly a very different work-flow compared to tweaking abstract hyper-parameters which implicitly affect the model performance. For example, if we were to work in the $\beta$-VAE setting, we would observe this transition: 
$\text{NLL} + \beta \text{KL} \Longrightarrow \text{KL} + \beta \text{RE}_\kappa$, where $\text{RE}_\kappa$ is the reconstruction error constraint as defined in \Cref{table.constraints}, and $\kappa$ is a tolerance hyper-parameter.
On the lhs $\beta$ is an hyper parameter tuning the relative weight of the negative log-likelihood (NLL) and KL terms, affecting model reconstructions in a non-smooth way as shown in \Cref{fig.phasetransition} and, as discussed in \Cref{sec.kernels}, implicitly defining a constraint on the VAE reconstruction error. On the rhs $\beta$ is a Lagrange multiplier, whose final value is automatically tuned during optimization as a function of the $\kappa$ tolerance hyper-parameter, which the user can define in pixel space explicitly specifying the required reconstruction performance of the model. 

\begin{algorithm}[H]
 \KwResult{Learned parameters $\theta$, $\eta$ and Lagrange multipliers $\vb$}
 Initialize $t=0$\;
 Initialize $\vb=\mathbf{1}$\;
 \While{is training}{
  Read current data batch $\vx$\;
  Sample from variational posterior $\vz \sim q(\vz | \vx)$\;
  Compute the batch average of the constraint $\hat{C}^t \leftarrow \mathcal{C}(\vx^t,g(\vz^t))$\;
  \eIf{$t == 0$}{
    Initialize the constraint moving average $C_{ma}^0 \leftarrow \hat{C}^0$\;
    }
    {
    $C_{ma}^{t} \leftarrow \alpha C_{ma}^{t-1} + (1-\alpha)\hat{C}^t$\;
    }
  $C^t \leftarrow \hat{C}^t + \text{StopGradient}(C_{ma}^{t}-\hat{C}^t)$\;
  Compute gradients $G_{\theta} \leftarrow \frac{\partial \mathcal{L}_{\vb}}{\partial \theta }$ and $G_{\eta} \leftarrow \frac{\partial \mathcal{L}_{\vb}}{\partial \eta }$\;
  Update parameters as $\Delta_{\theta, \eta} \propto - G_{\theta, \eta}$ and Lagrange multiplier(s) $\Delta_{\log(\vb)} \propto C^{t}$\;
%   Update parameters as $\Delta_{\theta, \eta} \propto - G_{\theta, \eta}$\;
%   Update Lagrange multiplier(s) ${\vb^t} \leftarrow $\vb^{t-1}\exp({\propto C^{t}})$\;
  $t \leftarrow t+1$\;
 }
 \caption{\badaptshort{}. Pseudo-code for joint optimization of VAE parameters and Lagrange multipliers. The update of the Lagrange multipliers is of the form ${\vb^t} \leftarrow \vb^{t-1}\exp({\propto C^{t}})$; this to enforce positivity of $\vb$, a necessary condition \cite{bertsekas1996} for tackling the inequality constraints. The parameter $\alpha$ controls the slowness of the moving average, which provides an approximation to the expectation of the constraint.}  \label{alg.beta}
\end{algorithm}

%% file: experiments.tex
\section{Experiments} \label{sec.experiments}

\subsection{Constrained optimization for large scale models and larger datasets} \label{sec.experiments.large_models}

We demonstrate empirically that \badaptshort{} provides an easy and robust way of balancing compression versus different types of reconstruction. We conduct experiments using standard implementations of ConvDraw \cite{Gregor2016TowardsCC} (both in the conditional and unconditional case) and a VAE+NVP model that uses a convolutional decoder similar to \cite{Rosca2018DistributionMI} and a fully connected conditional NVP \cite{dinh2016density} model as the encoder density so that we can approximate high-capacity encoders.

In \Cref{table.constraints} we show a few examples of reconstruction constraints that we have considered in this study. To inspect the performance of \badaptshort{} we look specifically at the behavior of trained models in the information plane (negative reconstruction likelihood vs KL) on various datasets, with and without the RE constraint. All models were trained using Adam \cite{kingma2014adam}, with learning rate of 1e-5 for ConvDraw and 1e-6 for the VAE+NVP, and a constraint moving average parameter $\alpha=0.99$. 

% \subsubsection{$\beta$-adaptation experiments on CelebA, Cifar10 and MNIST variants for unconditional and conditional models}

\begin{figure}[t]
\centering
\includegraphics[width=\textwidth]{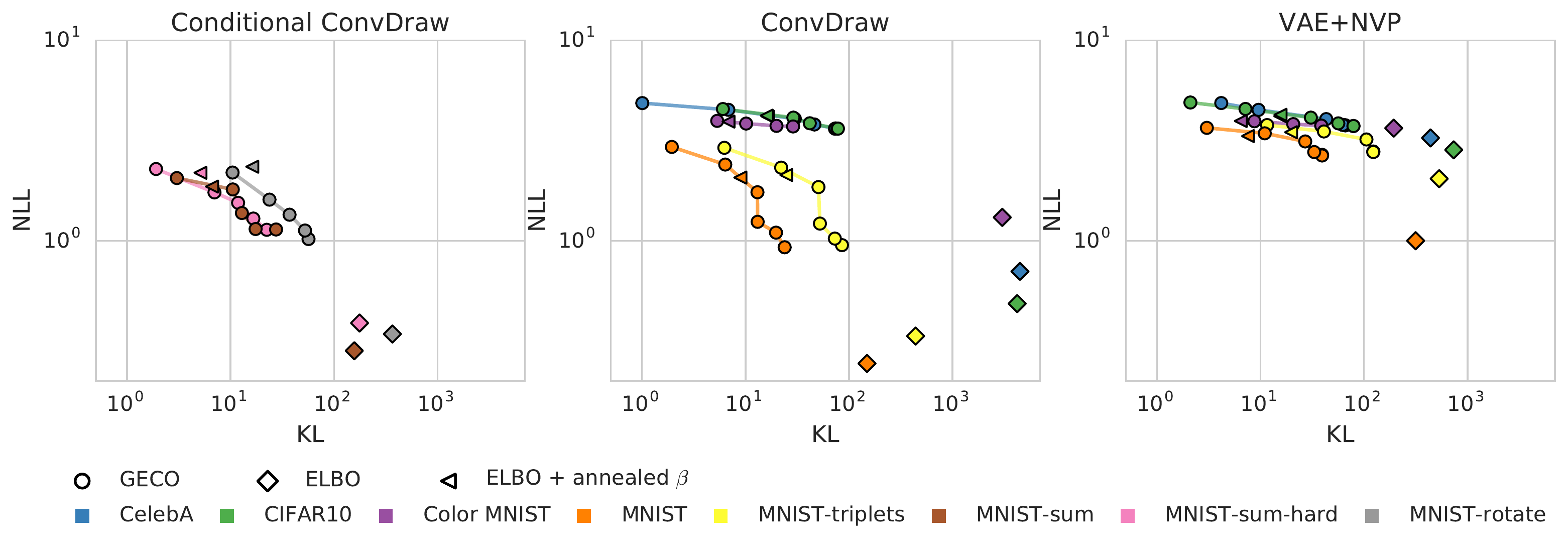}
\caption{
{\bf Information plane analysis of Conditional ConvDraw, ConvDraw and VAE+NVP with and without RE constraints}. 
Each plot shows the final reconstruction / compression trade-off achieved during training for the same ConvDraw and VAE+NVP models using ELBO, \badaptshort{} and ELBO with a hand annealed $\beta$, respectively. For \badaptshort{} we report results for the following reconstruction thresholds $\kappa \in \{0.06, 0.08, 0.1, 0.125, 0.175\}$, and visually tie them together by connecting them via a line colour-coded by the dataset instance they refer to. For the hand annealed $\beta$ we use the same annealing scheme reported in \cite{GQN}. Results are shown for a variety of conditional and unconditional datasets, providing evidence of the consistency of the behavior of \badaptshort{} across different domains.   
}
\label{fig.large_experiments}
\end{figure}

\begin{table}[h]
\centering
  \begin{tabular}{|c|c|}
  \hline
    {{\bf Name}} & {$\mathcal{C} (x, g (z))$}\\ \hline
    Reconstruction Error (\textbf{RE}) & $\|x-g(x)\|^2 - \kappa^2$\\ %\hline
    Feature Reconstruction Error (\textbf{FRE}) & $\|f(x)-f(g(x)) \|^2 - \kappa^2$\\ %\hline
    Classification accuracy (\textbf{CLA}) & $l(x)^T c (g (z)) - \kappa$\\ 
    Patch Normalized Cross-correlation (\textbf{pNCC}) & $\sum\left[\kappa - \psi(x, i) ^T \psi(g(x), i)\right]$\\ \hline
  \end{tabular}
  \vspace{1mm}
  \caption{Constraints studied in this paper. For the FRE constraint, the features $f(\vx)$ can be extracted a classification network (in our experiments we use a pretraiend Cifar10 resnet), or compute local image statistics, such as mean and standard deviation. For the CLA constraint, $c(\vx)$ is a simple convolutional MNIST classifier that outputs class probabilities and $l(\vx)$ is the one-hot true label vector of image $\vx$. For the pNCC constraint we define the operator $\psi(x, i)$, which returns a whitened fixed size patch from input image $x$ at location $i$, and constraint the dot products of corresponding patches from targets and reconstructions.} \label{table.constraints}
\end{table}

Here we look at the behavior of VAE+NVP and ConvDraw (the latter both in the conditional and unconditional case) in information plane (negative reconstruction likelihood vs KL) on various datasets, with and without a RE constraint.

The datasets we use for the unconditional case are CelebA\cite{liu2015faceattributes}, Cifar10\cite{krizhevsky2009learning}, MNIST\cite{lecun2010mnist}, Color-MNIST\cite{unrolledgan} and a variant of MNIST we will refer to as MNIST-triplets. MNIST-triplets is comprised of triplets of MNIST digits $\{(\vec{I}_i, l_i)\}_{i=0,1,2}$ such that $l_2 = (l_0 + l_1) \bmod 10$; the model is trained to capture the joint distribution of the image vectors $\{(I_{i,0}, I_{i,1}, I_{i,2})\}_i$.

In the conditional case we use are variants of MNIST we will refer to as MNIST-sum, MNIST-sum-hard and MNIST-rotate. All variants of the datasets are comprised of contexts and targets derived from triplets of MNIST digits $\{(\vec{I}_i, l_i)\}_{i=0,1,2}$, with constraints as follows.
For MNIST-sum contexts are $\{(I_{i,0}, I_{i,1})\}_i$ and targets are $\{I_{i,2}\}_i$, such that  $l_2 = (l_0 + l_1) \bmod 10$;
for MNIST-sum-hard contexts are $\{I_{i,2}\}_i$ and targets are $\{(I_{i,0}, I_{i,1})\}_i$, such that  $l_2 = (l_0 + l_1) \bmod 10$;
finally, for MNIST-rotate contexts are $\{(I_{i,0}, I_{i,1})\}_i$ and targets are $\{\hat{I}_{i,2}\}_i$, such that $l_2 = l_0$ and $\hat{I_2}$ is $I_2$ rotated about its centre by $l_1 \cdot 30^\circ$, note that whilst $I_0$ and $I_2$ have the same label, they are not the same digit instance.

In figure \Cref{fig.large_experiments} we show in all cases that the negative reconstruction log-likelihoods (NLL, see \Cref{sec.nll} for details) reached by VAE+NVP, ConvDraw and conditional ConvDraw trained only with the ELBO objective are lower compared to the values obtained with ELBO + \badaptshort{}, at the expense of KL-divergences that are some times many orders of magnitude higher. 
This result comes from the observation that the numerical values of reconstruction errors necessary to achieve good reconstructions can be much larger, allowing the model to achieve lower compression rates. To provide a notion of the quality of the reconstructions when using \badaptshort{}, we show in \Cref{fig.samples.draw.teaser} a few model samples and reconstructions for different reconstruction thresholds and constraints. In \Cref{sec.samples} we show reconstructions and samples for all levels of reconstruction targets.
As we can see from \Cref{fig.samples.draw.teaser}, the use of different constraints has a dramatic impact on both the quality of reconstructions and samples.

\subsubsection{Average and Marginal KL analysis}
\label{sec.marginal_kl}

\begin{figure}[!ht] 
\centering
\includegraphics[width=0.8\textwidth]{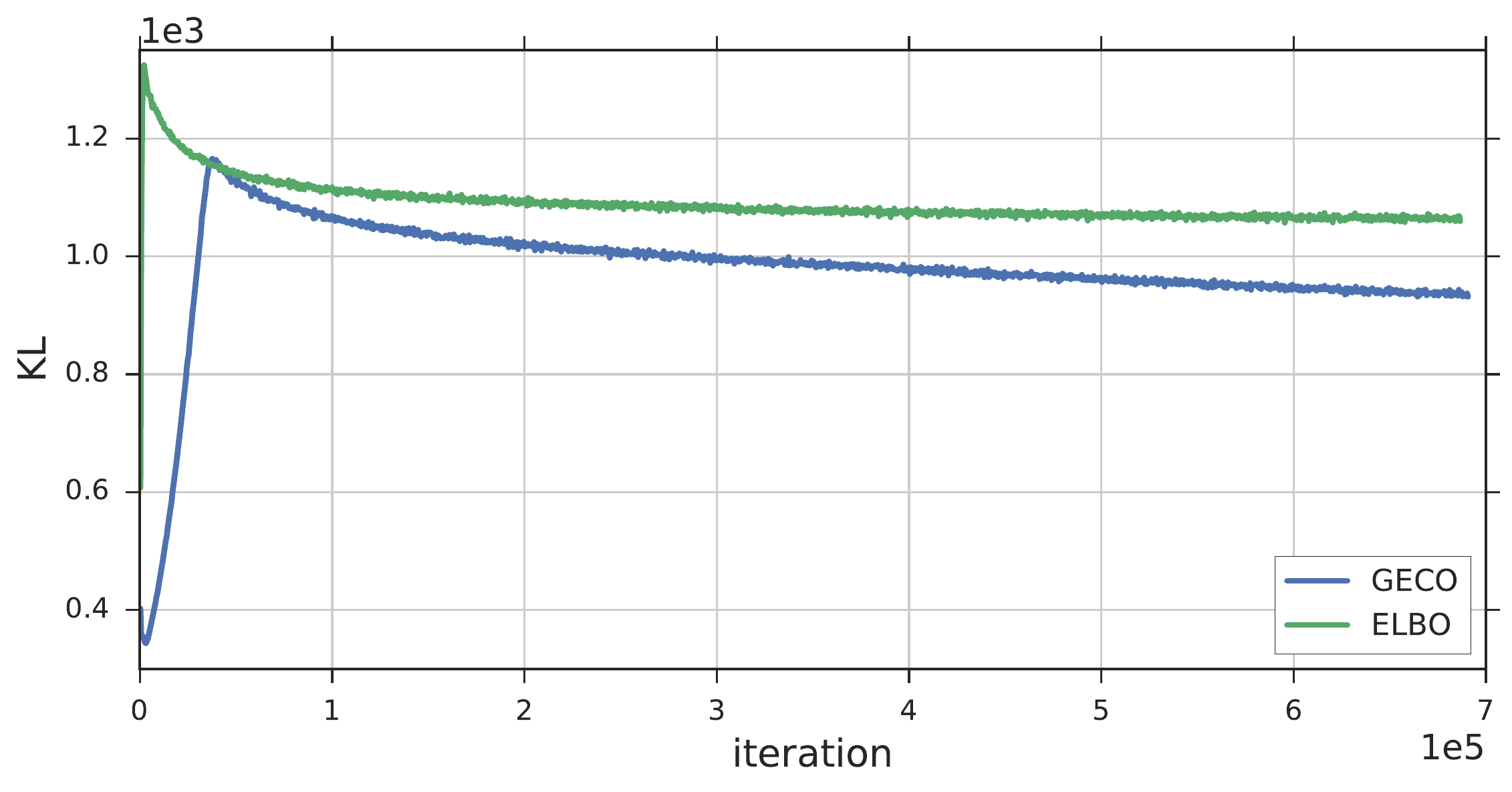}
\caption{
{\bf GECO results in lower average KL at fixed reconstruction error compared to ELBO.} We first trained an expressive ConvDRAW model on CIFAR10 using the standard ELBO objective until convergence and recorded its reconstruction error (MSE=0.00029). At this reconstruction error values, the reconstructions are visually perfect.
We then trained the same model architecture using GECO with a RE constraint setup to achieve the same reconstruction error. The curves for the model trained with ELBO (green) and with GECO (blue) demonstrate that we can achieve the same reconstruction error but with a lower average KL between prior and posterior.
}
\label{fig.elbo_vs_geco}
\end{figure}

At a fixed reconstruction error, a computationally cheap indicator of the quality of the learned encoder is the average KL between prior and posterior, $\frac{1}{n}\sum_i \KL{q(\vz| \vx_i)}{\pi(\vz)}$, which we analyze in \Cref{fig.elbo_vs_geco}. Our analysis shows that an expressive model can achieve lower average KL at a given reconstruction error when trained with GECO compared to the same model trained with ELBO.

From \Cref{prop.tiling,prop.equipartition}, the optimal solutions for VAE's encoders are inference models that cover the latent space in such a way that their marginal is equal to the prior. That is, $q(\vz) = \frac{1}{n}\sum_i q(\vz| \vx_i) = \pi(\vz)$. We refer to the KL between $q(\vz)$ and $\pi(\vz)$ as the "marginal KL".

If the learned encoder or inference network fails to cover the latent space, it may result in the "holes" problem which, in turn, is associated with bad sample quality.

In contrast to the average KL, the marginal KL is also sensitive to the "holes problem" discussed in \Cref{sec.intro}.
In \Cref{table.marginal_kl} we evaluate the effect of \badaptshort{} on the marginal KL of the VAE+NVP models trained in \Cref{sec.experiments.large_models} (in a limited number of combinations due to the significant computational costs) and observe that models trained with \badaptshort{} also have much lower marginal KL while maintaining an acceptable reconstruction accuracy.

\begin{table}[h]
\centering
  \begin{tabular}{|c|c|c|}
  \hline
    {{\bf Dataset}} & {\bf Marginal KL for ELBO} & {\bf Marginal KL for \badaptshort{}}  \\ \hline
    Cifar10 & 725.2 & 45.3 \\ \hline
    Color-MNIST & 182.5 & 10.3 \\ \hline
  \end{tabular}
  \vspace{2mm}
  \caption{Marginal KL comparison for a VAE+NVP model on Cifar10 and Color-MNIST.} \label{table.marginal_kl}
\end{table}

%% file: discussion.tex
\section{Discussion}
\begin{figure}[t] 
\centering
\includegraphics[width=\textwidth]{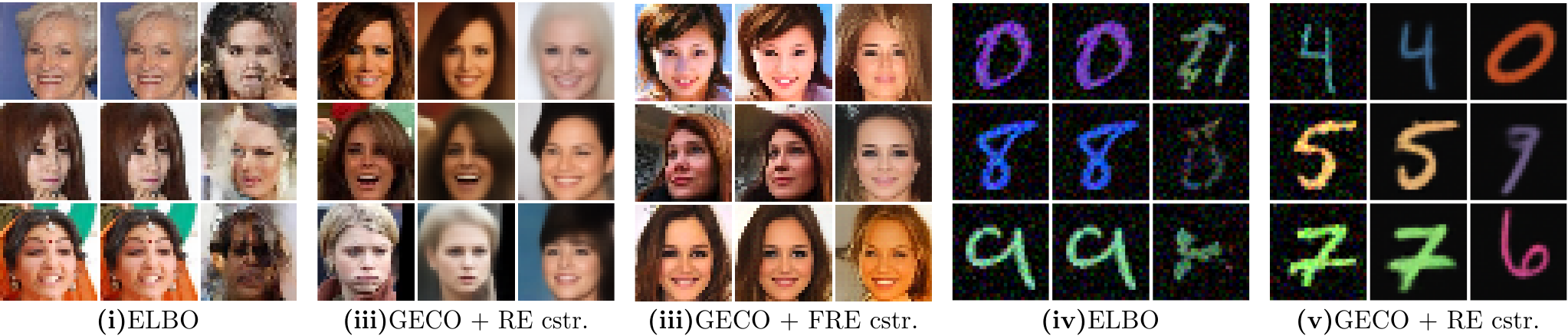}
\caption{
{\bf Examples of samples and reconstructions from ConvDraw trained on CelebA and Color-MNIST}. In each block of samples, rows correspond to samples from the data, model reconstructions and model samples respectively. From left to right we have models trained with: 
(i) ELBO only.
(ii) ELBO + \badaptshort{}+RE constraint with $\kappa=0.1$.
(iii) ELBO + \badaptshort{}+FRE constraint with $\kappa=1.0$.
(iv) ELBO only. 
(v) ELBO + \badaptshort{}+RE constraint with $\kappa=0.06$.
More samples available in \Cref{sec.samples}.
}
\label{fig.samples.draw.teaser}
\end{figure}

We have provided a detailed theoretical analysis of the behavior of high-capacity VAEs and variants of $\beta$-VAEs. We have made connections between VAEs, spectral clustering methods and statistical mechanics (phase transitions). Our analysis provides novel insights to the two most common problems with VAEs: blurred reconstructions/samples, and the "holes problem".
Finally, we have introduced \badaptshort{}, a simple-to-use algorithm for constrained optimization of VAEs. Our experiments indicate that it is a highly effective tool to achieve a good balance between reconstructions and compression in practice (without recurring to large parameter sweeps) in a broad variety of tasks.

\section*{Acknowledgements}

We would like to thank Mihaela Rosca for the many useful discussions and the help with the Marginal KL evaluation experiments.

%% file: appendix.tex
\section{Reconstruction fixed-points experiment details} \label{sec.experiments.fixedpoints}

To produce the results in \Cref{fig.phasetransition}, we iterated the fixed-point equations for the matrix $\psi^t$ using exponential smoothing with $\alpha = 0.9$. For each experiment, we iterated the smoothed fixed-point equations until either the number of iterations exceeded 400 steps or the Euclidean distance between two successive steps became smaller than 1e-3.
The basis functions $\phi_i(z)$ were arranged as a 32x32 grid in a compact latent space $\vz \in [-\frac{1}{2},\frac{1}{2}]^2$, the prior was chosen to be uniform, $\pi(\vz)=1$. The matrix $\psi$ was initialized at the center-points of the grid tiles with small uniform noise in $[-0.1, 0.1]$.

\section{High-capacity $\beta$-VAEs and Lipschitz constraints}\label{sec.lipschitz}

The analysis of high-capacity VAEs in \Cref{sec.tiles,sec.kernels} reveal interesting aspects of VAEs near convergence, but the type of solutions implied by \Cref{prop.tiling,prop.kernel} may seem unrealistic for VAEs with smooth decoders parametrized by deep neural networks. For instance, the solutions of the fixed point-equations from \Cref{prop.tiling,prop.kernel} have no notion that the outputs of the decoder $g(\vz)$ and $g(\vz')$ for two similar latent vectors $\vz$ and $\vz'$ should also be similar. That is, these solutions are not sensitive to the metric and topological properties of the latent space.
This implies that, if we want to work in a more realistic high-capacity regime, we must further constrain the solutions so that they have at least a continuous limit as we grow the number of basis functions to infinity. 

A sufficient condition for a function $g(\vz)$ to be continuous, is that it is a locally-Lipschitz function of $\vz$ \cite{o2006metric}. Thus, to bring our analysis closer to more realistic VAEs, we consider high-capacity $\beta$-VAEs with an extra $L$-Lipschitz inequality constraint $\mathcal{C}$ in the decoder function $g(\vz)$. The new term that we add to the augmented Lagrangian, with a functional Lagrange multiplier $\Omega(\vz, \vz') \ge 0$ is given by
\begin{align}
    \mathcal{C}[g] &= \frac{1}{2}\int d\vz d\vz' \pi(\vz) \pi(\vz') \Omega(\vz, \vz') \left[ \| g(\vz) - g(\vz') \|^2 - L^2 \| \vz - \vz' \|^2 \right]. \label{eq.lipschitz}
\end{align}
By expressing $g(\vz)$ and $\Omega(\vz, \vz')$ in the functional basis $\phi(\vz)$ of \Cref{prop.kernel}, $g(\vz) = \sum_a \psi_a \phi_a(\vz)$ and $\Omega(\vz, \vz') = \sum_{a, b} \hat{\Omega}_{a b}\phi_a(\vz)\phi_b(\vz)$, we can rewrite the constraint term as a quadratic inequality constraint in the matrix $\psi$,
\begin{align}
    \mathcal{C}[g] &= \frac{1}{2}\sum_{a, b} \tilde{\Omega}_{a b} \left[ C_{a b} \| \psi_a - \psi_b \|^2 - 1 \right], \label{eq.lipschitz.psi}
\end{align}
where $\tilde{\Omega}_{a b} =  L^2 K_{a b} \hat{\Omega}_{a b}$ are new Lagrange multipliers, $C_{a b} = \pi_a \pi_b / (L^2 K_{a b})$ and $K_{a b} = \int d\vz d\vz' \phi_a(\vz) \phi_b(\vz')\| \vz - \vz' \|^2$. The matrices $C_{a b}$ and $K_{a b}$ embed the metric and topological properties of the latent space but are otherwise independent of the rest of the model.

The constraint \eqref{eq.lipschitz} can be used to enforce both global and local Lipschitz constraints by controlling the size of the support of the Lagrange multiplier function $\Omega(\vz, \vz')$. If $\Omega(\vz, \vz') = 0$ for $\| \vz - \vz' \| >= r$, we will be constraining the decoder function to be locally Lipschitz within a radius $r$ in the latent space.

\begin{figure}[t] 
\centering
\includegraphics[width=\textwidth]{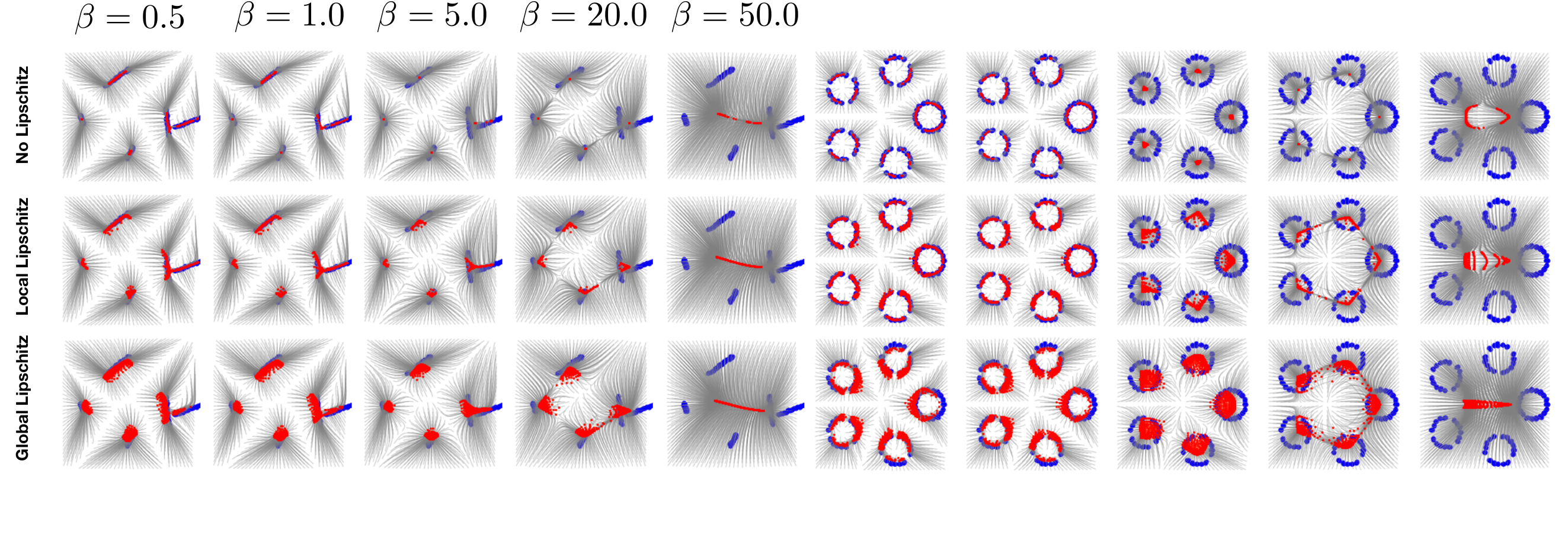}
\caption{
{\bf Combined effect of $\beta$-VAEs and Lipschitz constraints}. Blue points are data-points for a mixture of lines (left) and a mixture of circles (right). Grey curves indicate the trajectories of the reconstruction vectors $\psi$ from initial conditions on an uniform 2D grid. Red points are the found fixed-points. {\bf Top row} Reconstruction fixed-points without Lipschitz constraints. An increase in $\beta$ factor causes the reconstruction fixed-points to collapse, effectively clustering the data at different spatial resolutions; {\bf Middle row} Reconstruction fixed-points with local Lipschitz constraints ($r=0.2$). As we increase the strength of local Lipschitz constraints, the fixed-points tend organize in thin manifolds connecting regions of high density.
{\bf Bottom row} Reconstruction fixed-points with global Lipschitz constraints. As we increase the strength of global Lipschitz constraints ($r=1.0$), the fixed-points tend organize in manifolds covering regions of high density.
}
\label{fig.lipschitz}
\end{figure}

Note that the Lipschitz constraint is not a reconstruction constraint as it only constrains the VAE's decoder at arbitrary points in the latent space. For this reason, it can be implemented as a projection step just after the iteration from \Cref{eq.fp.beta.psi}. This is formalized in \Cref{prop.lipschitz.psi}. We illustrate the combined effect of $\beta$, local and global Lipschitz constraints on VAEs in \Cref{fig.lipschitz}.

\begin{proposition} \label{prop.lipschitz.psi}
  The Lipschitz constraint from \Cref{eq.lipschitz.psi} can be incorporated to the fixed-point \Cref{eq.fp.beta.psi} as a projection of the form $\psi^{t+1} = F(\psi^{t}) P^t$, where $F$ is the transition operator without Lipschitz constraints.
  See proof in \Cref{proof.lipschitz.psi}.
\end{proposition}

\section{Proofs}

\subsection{Derivation of \Cref{prop.fixedpoint.eqs}}
\begin{proof}\label{proof.fixedpoint.eqs}
    We can obtain these equations by taking the functional derivatives
    $\frac{\delta \mathcal{F}}{\delta g(\vz)}$, $\frac{\delta \mathcal{F}}{\delta q(\vz | \vx)}$, $\frac{\delta \mathcal{L}_{\vb}}{\delta g(\vz)}$ and $\frac{\delta \mathcal{L}_{\vb}}{\delta q(\vz | \vx)}$ of $\mathcal{F}$ and $\mathcal{L}_{\vb}$ with respect to $g(\vz)$ and $q(\vz | \vx)$ respectively and re-arranging the terms of the equations $\frac{\delta \mathcal{F}}{\delta q(\vz | \vx)}=0$, $\frac{\delta \mathcal{F}}{\delta g(\vz)}=0$, $\frac{\delta \mathcal{L}_{\vb}}{\delta q(\vz | \vx)}=0$ and $\frac{\delta \mathcal{L}_{\vb}}{\delta g(\vz)}=0$. For the density $q(\vz | \vx)$ we must also take the normalization constraint $\int d\vx d\vz \lambda (\vx) (q(\vz | \vx)-1)$ into account:
\begin{align}
\frac{\delta \mathcal{F}}{\delta g(\vz)} &= \frac{1}{\sigma^2} \left( \EE{\rho(\vx)}{q(\vz | \vx)\vx} - \EE{\rho(\vx)}{q(\vz | \vx)}g(\vz)\right) = 0, \label{eq.grads.g.elbo}\\
\frac{\delta \mathcal{F}}{\delta q(\vz | \vx)} &= \rho(\vx)\left[ \frac{ \| \vx - g(\vz) \|^2 }{2 \sigma^2}  + \ln \frac{q(\vz | \vx)}{\pi(\vz)} + 1 + \lambda(\vx) \right] = 0, \label{eq.grads.q.elbo}\\
\frac{\delta \mathcal{L}_{\vb}}{\delta g(\vz)} &= - \vb^T\EE{\rho(\vx)}{ q (\vz |  \vx) \frac{\partial \mathcal{C}( \vx, g(\vz))}{\partial g(\vz)} } = 0, \label{eq.grads.g}\\
\frac{\delta \mathcal{L}_{\vb}}{\delta q(\vz | \vx)} &= \rho(\vx)\left[ \ln \frac{q(\vz | \vx)}{\pi(\vz)} + 1 + \vb^T\mathcal{C}( \vx, g(\vz))  + \lambda(\vx) \right] = 0, \label{eq.grads.q}
\end{align}
where $\mathcal{F}$ was computed using a Gaussian likelihood with global variance $\sigma^2$.
A straightforward algebraic simplification of these equations results in the fixed-point equations of the \Cref{prop.fixedpoint.eqs}. 
For a general constraint $\mathcal{C}(\vx,g(\vz))$ we cannot simply solve $\frac{\delta \mathcal{L}_{\vb}}{\delta g(\vz)}=0$ for $g(\vz)$ due to the non-linear dependency of $\mathcal{C}(\vx,g(\vz))$ on $g(\vz)$. In this case, we can often employ a standard technique for constructing fixed-point equations which consists in converting an equation of the form $x - f(x)=0$ to a recurrence relation of the form $x_n = f(x_{n-1})$. 
\end{proof}
  
\subsection{Derivation of \Cref{prop.tiling}}
\begin{proof}\label{proof.tiling}
    First, we note that for any given $q (\vz | x)$, the ELBO is a convex quadratic functional of the decoder $g(\vz)$ and, for a fixed $g(\vz)$, it is a convex functional of the encoder $q (\vz | x)$. Second, for a fixed partition $\Omega_i$, replacing the solution $q^t (\vz |  x_i) =  \pi (\vz) \mathbb{I}_{z
    \in \Omega_i}/\pi_i$, where $\pi_i = \EE{\pi}{\mathbb{I}_{z
    \in \Omega_i}}$, in the fixed-point equations, results in itself. That is, 
    \begin{align}
      w_i^t (\vz) &= \frac{q^t (\vz |  x_i)}{\sum_j q^t (\vz | \vx_j)} 
      = \mathbb{I}_{z \in \Omega_i}\\
      g^t (\vz) &= \sum_i w_i^t (\vz) x_i = \sum_i x_i \mathbb{I}_{z \in
      \Omega_i}\\
      \sigma^t &= \sqrt{\mathbb{E}_{\rho (\vx) q^t (\vz | \vx )} [\| \vx - g^t(\vz) \|^2]} = 0\\
      q^{t + 1} (\vz |  x_i) &= \lim_{\sigma \rightarrow 0} \frac{\pi (\vz)
      e^{- \frac{\| x_i - g^t (\vz) \|^2}{2 \sigma^2}}}{c (x_i)} = \lim_{\sigma
      \rightarrow 0} \frac{\pi (\vz) \sum_j e^{- \frac{\| x_i - x_j \|^2}{2
      \sigma^2}} \mathbb{I}_{z \in \Omega_j}}{\int d z' \pi (\vz') \sum_k e^{-
      \frac{\| x_i - x_k \|^2}{2 \sigma^2}} \mathbb{I}_{z \in \Omega_k}} = \frac{1}{\pi_i} \pi
      (\vz) \mathbb{I}_{z \in \Omega_i}.
    \end{align}
    Therefore, $q^t (\vz |  x_i) =  \pi (\vz) \mathbb{I}_{z
    \in \Omega_i}/\pi_i$ is a fixed-point in the family of densities constrained by the partition $\Omega_i$. We observe that the negative ELBO reduces to the expected KL term only, $\mathbb{E}_p [\text{KL} (q ; \pi)] = \frac{1}{n} \sum_j \int d z 
    \frac{\pi (\vz) \mathbb{I}_{z \in \Omega_j}}{\pi_j}  (-\ln \pi_j) = -\frac{1}{n} \sum_j \ln \pi_j$.
    We can now optimize the partition to further maximize the ELBO. This results in $\pi_j=1/n$. That is, the tiles $\Omega_i$ must be equiprobable.
    At this point, we have $\mathbb{E}_p [\text{KL} (q ; \pi)] =\ln n$.
\end{proof}

\subsection{Derivation of \Cref{prop.blurredrecs}}
\begin{proof} \label{proof.blurredrecs}
From \eqref{eq.fp.g} we have that $g(\vz) = \sum_i w_i^t (\vz) \vx_i = \sum_i \vx_i \frac{q (\vz |  x_i)}{\sum_j q (\vz | \vx_j)}$. Substituting $q(\vz | x_i)=\pi (\vz) \mathbb{I}_{z \in \Omega_i}/\pi_i$ we have,
\begin{align}
    g(\vz) &= \sum_i \vx_i \frac{q (\vz |  x_i)}{\sum_j q (\vz | \vx_j)} = \sum_i \vx_i \frac{\mathbb{I}_{z \in \Omega_i}/\pi_i}{\sum_j \mathbb{I}_{z \in \Omega_j}/\pi_j} = \sum_{i | \vz \in \Omega_i} \vx_i \frac{1/\pi_i}{\sum_{j | \vz \in \Omega_j} 1/\pi_j},
\end{align}
where $\pi_i = \EE{\pi}{\mathbb{I}_{z \in \Omega_i}}$.
\end{proof}

\subsection{Derivation of \Cref{prop.kernel}}
\begin{proof} \label{proof.kernel}
    The expression for the generator can be obtained by substitution and
    algebraic simplification using the fact that $\phi_a$ form an orthogonal
    basis and that $\phi_a (\vz) \in \{ 0, 1 \}$:
    \begin{align}
      g (\vz) &= \sum_{i b} \frac{q^t (\vz |  \vx_i)}{\sum_j q^t (\vz |
       \vx_j)} \vx_i = \sum_{i a} \frac{m_{i a} \phi_a (\vz)}{\sum_{j b} m_{j b} \phi_b (\vz)}
      \vx_i = \sum_{i b} \frac{m_{i b}}{\sum_j m_{j b}} \phi_b (\vz) \vx_i = \psi^T \phi (\vz),
    \end{align}
    with $\psi_b = \sum_i \frac{m_{i b}}{\sum_j m_{j b}} \vx_i$. Similarly, we
    can compute the fixed point equations for $\psi_a$ by substitution on
    equations \eqref{eq.fp.q} and \eqref{eq.fp.g},
    \begin{align}
      q_i^{t + 1} (\vz) &= \pi (\vz) \sum_b \frac{e^{- \frac{\| x_i - \psi^t_b
      \|^2}{2 \beta}}}{c^t_i} \phi_b (\vz) = \pi (\vz) \sum_a m_{i a}^{t + 1}
      \phi_a (\vz) \\
      \psi^{t + 1}_b &= \sum_i x_i \frac{m^{t + 1}_{i b}}{\sum_j m^{t + 1}_{j b}} = \frac{\sum_i x_i \frac{e^{- \frac{\| x_i - \psi^t_b \|^2}{2
      \beta}}}{c^t_i}}{\sum_i \frac{e^{- \frac{\| x_i - \psi^t_b \|^2}{2
      \beta}}}{c^t_i}}, \label{eq.fp.beta.psi}
    \end{align}
    where $c_i = \sum_b e^{- \frac{\| x_i - \psi^t_b \|^2}{2 \beta}} \pi_b$ and $m^{t + 1}_{i b} = \frac{e^{- \frac{\| x_i - \psi^t_b \|^2}{2
      \beta}}}{c^t_i}$.
    If the initial reconstruction vectors $\psi_a^t$ are in the convex-hull of the training data-points, then equations \eqref{eq.fp.beta.psi} will map them to another set of points $\psi_a^{t+1}$ in the convex-hull of the training data. Since, these equations are also smooth with respect to $\psi$, they are guaranteed to converge as a consequence of the fixed-point theorem \cite{rosenlicht1968introduction}.
    Importantly, equation \eqref{eq.fp.beta.psi} corresponds to the fixed point iterations for computing the pre-image (reconstructions) of Kernel-PCA but using a normalized Gaussian kernel. 
  \end{proof}

\subsection{Derivation of \Cref{prop.lipschitz.psi}}
\begin{proof} \label{proof.lipschitz.psi}
   We first write the relevant terms of the augmented Lagrangian $\mathcal{L}_{\vb, \Omega}$ in the basis $\phi(\vz)$ from \Cref{prop.kernel},
  \begin{align}
    \mathcal{L}_{\vb, \Omega} &= \EE{\rho(\vx)}{\EE{q (\vz |  \vx)}{ \frac{ \| \vx - \psi^T\phi(\vz) \|^2 }{2 \sigma^2}}} + \frac{1}{2n}\sum_{a, b} \tilde{\Omega}_{a b} \left[ C_{a b} \| \psi_a - \psi_b \|^2 - 1 \right] + \text{cst}\\
    &= \EE{\rho(\vx)}{\EE{q(\vz|\vx)}{\phi_a(\vz)} \frac{ \| \vx - \psi_a \|^2 }{2 \sigma^2} }  + \frac{1}{2n}\sum_{a, b} \tilde{\Omega}_{a b} \left[ C_{a b} \| \psi_a - \psi_b \|^2 - 1 \right] + \text{cst}\\
    &= \frac{1}{n}\sum_{i a} m_{i a}\pi_a \frac{ \| \vx_i - \psi_a \|^2 }{2 \sigma^2}  + \frac{1}{2n}\sum_{a, b} \tilde{\Omega}_{a b} C_{a b} \| \psi_a - \psi_b \|^2 + \text{cst},
 \end{align}
where cst are terms that do not depend on $\psi$ for a fixed $q$. Solving $\frac{\partial \mathcal{L}_{\vb, \Omega}}{\partial \psi_a}=0$ with respect to $\psi$ results in 
    \begin{align}
    \frac{\partial \mathcal{L}_{\vb, \Omega}}{\partial \psi_a} &= \frac{\pi_a \sum_i m_{i a}}{\sigma^2} \left [ \sum_i \vx_i \frac{m_{i a}}{\sum_i m_{i a}} -  \psi_a  + \frac{\sigma^2}{\pi_a \sum_i m_{i a}} \sum_{b} \tilde{\Omega}_{a b} C_{a b} ( \psi_a - \psi_b ) \right]=0\\
    \psi_b &= \sum_{i, a} \vx_i \frac{m_{i a}}{\sum_i m_{i a}} P_{a b},
    \end{align}
    where $P = [\id - \text{diag}(\frac{n \sigma^2}{1^T m\bigodot\pi}) ( \text{diag}(1^T (\tilde{\Omega} \bigodot C)) - \tilde{\Omega} \bigodot C ) ]^{-1}.$
\end{proof}

\subsection{Derivation of \Cref{prop.equipartition}}
\begin{proof} \label{proof.equipartition}
From \Cref{prop.kernel}, we have seen that the reconstruction vectors $\psi_a$ of a high-capacity $\beta$-VAE will 
converge to a set of $m$ fixed-points. This means that $\psi$ will map the set basis-functions to a smaller subset of points. As a consequence, all the latent-vectors falling in the support of these basis elements 
will also map to the same reconstruction and for a fixed $\vx$, the function $H(\vx, \vz)$ will only have $m$ possible distinct values, which we can enumerate as $H_{i a}$. 
If we enumerate all distinct supports as $\Omega_a$, 
then from \Cref{eq.fp.q} we have that $q_i(\vz) = \pi(\vz) \sum_a \frac{e^{- \frac{H_{i a}}{\beta}}}{\sum_b e^{-
\frac{H_{i b}}{\beta}} \gamma_b}  \mathbb{I}_{\vz \in \Omega_a}$.
Replacing $q_i(\vz)$ in \Cref{eq.lagrangian} and maximizing with respect to $\gamma_a$ results in the condition $\sum_i \frac{e^{- \frac{H_{i a}}{\beta}}}{\sum_b e^{- \frac{H_{i
b}}{\beta}} \gamma_b}=n$.
\end{proof}

\section{Computing $\beta$-independent NLLs}\label{sec.nll}

When we train $\beta$-VAEs with different constraints using \badaptshort{}, it is not obvious how to compare them in the information plane due to the arbitrary scaling learned by the optimizer.
In order to do a more meaningful comparison after the models have been trained, we recompute an optimal global standard-deviation $\sigma_{\text{opt}}$ for all models on the training data (keeping all other parameters fixed):
\begin{align}
    \sigma_{\text{opt}} &\approx \sqrt{ \frac{1}{M} \sum_{i \in \text{training set}} \|\vx_i - g(\vz_i) \|^2 }
\end{align}
where and $\vz_i \sim q(\vz|\vx_i)$ and $M$ is the batch size used for this computation. All the reported negative reconstruction likelihoods were computed using $\sigma_{\text{opt}}$. We report all likelihoods per-pixel using the "quantized normal" distribution \cite{Rosca2018DistributionMI} to make it easier to compare with other models.

\clearpage
\section{Extra Experiments}
\label{appendix.extra_experiments}

\subsection{Micro-MNIST}

\begin{figure}[!ht] 
\centering
\includegraphics[width=\textwidth]{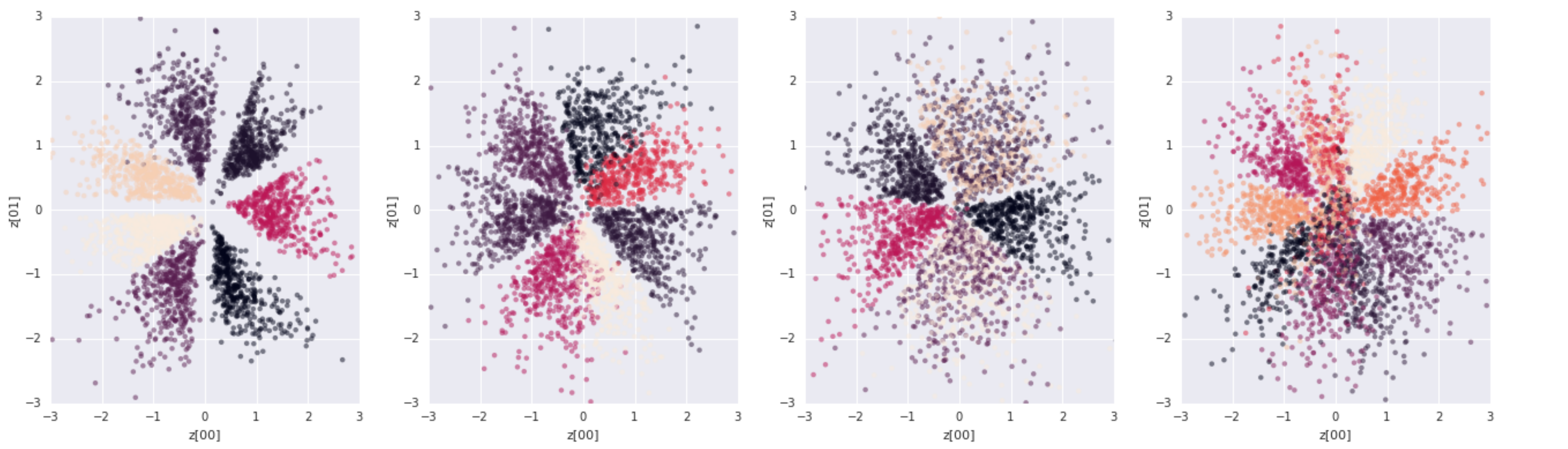}
\caption{
{\bf Learned posterior densities by a $\beta$-VAE+NVP via ELBO and with RE and CLA constraints on Micro-MNIST}. Each plot shows 100 samples from the learned encoder in a 2-dimensional latent space for each data-point from Micro-MNIST. The posterior samples are colored according to the data-point they belong to.
From left to right: 
(i) ELBO only $\beta=1$; 
(ii) $\beta$ optimized by \Cref{alg.beta} using a \badaptshort{}+RE constraint with $\kappa=0.05$;
(iii) $\beta$ optimized by \Cref{alg.beta} using a \badaptshort{}+RE constraint with $\kappa=0.5$;
(iv) $\beta$ optimized by \Cref{alg.beta} using a \badaptshort{}+CLA constraint with $\kappa=0.9$.
This experiment illustrates that VAEs with highly expressive posterior densities trained via ELBO have problems forming a tight tiling of the latent-space as predict by theory. By training the VAE+NVP with \badaptshort{} we can substantially tighten the posterior tiles.
}
\label{fig.micro_mnist}
\end{figure}

As a way to illustrate the impact of different constraints on the learned encoder densities, we created the "Micro-MNIST" dataset, comprised of only 7 random samples from MNIST.
In these experiments, shown in \Cref{fig.micro_mnist}, we observed that as we make the constraints weaker (larger $\kappa$ for the RE constraint or smaller $\kappa$ for the CLA constraint), we encourage the posterior density to collapse the modes of the data, focusing on the properties of the data relevant to the constraints.

\clearpage
\section{Model and Data Samples}\label{sec.samples}

\begin{figure}
    \centering
    \begin{subfigure}[t]{0.1\textwidth}
        \centering\includegraphics[width=1.2cm]{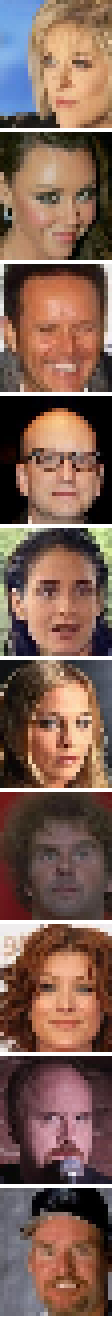}
        \caption{}
    \end{subfigure}
    \begin{subfigure}[t]{0.1\textwidth}
        \centering\includegraphics[width=1.2cm]{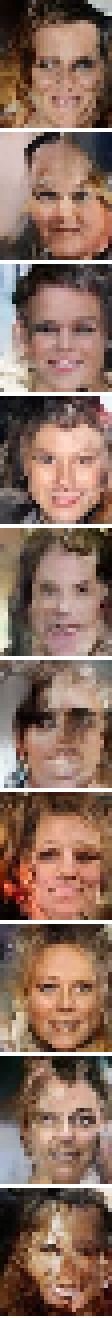}
        \caption{}
    \end{subfigure}
    \begin{subfigure}[t]{0.1\textwidth}
        \centering\includegraphics[width=1.2cm]{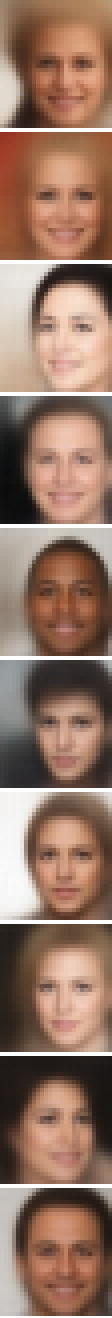}
        \caption{}
    \end{subfigure}
    \begin{subfigure}[t]{0.1\textwidth}
        \centering\includegraphics[width=1.2cm]{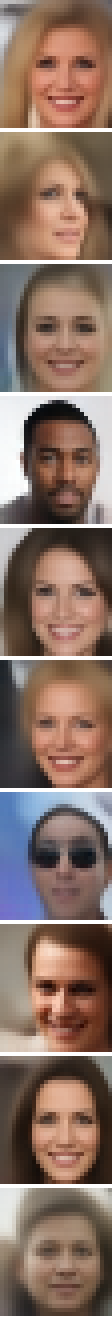}
        \caption{}
    \end{subfigure}
    \begin{subfigure}[t]{0.1\textwidth}
        \centering\includegraphics[width=1.2cm]{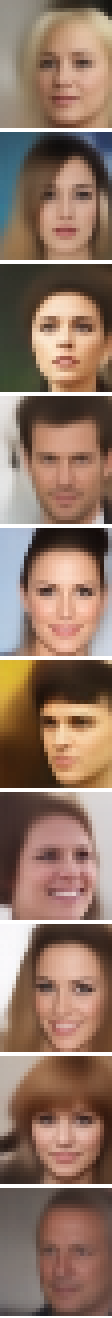}
        \caption{}
    \end{subfigure}
    \begin{subfigure}[t]{0.1\textwidth}
        \centering\includegraphics[width=1.2cm]{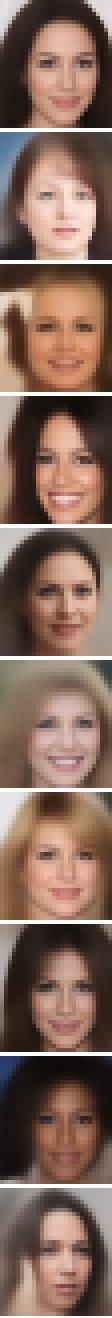}
       \caption{}
    \end{subfigure}
    \begin{subfigure}[t]{0.1\textwidth}
        \centering\includegraphics[width=1.2cm]{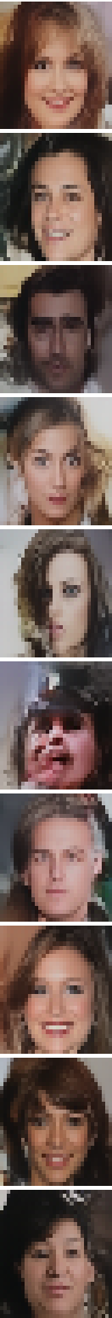}
       \caption{}
    \end{subfigure}

    \caption{
    {\bf Samples from ConvDraw trained on CelebA}. In each block of samples, rows correspond to samples from the data, model reconstructions and model samples respectively.
    From left to right we have models trained with: 
    (a) Data;
    (b) ELBO only; 
    (c) ELBO + Hand crafted $\beta \in [2.0, 0.7]$ annealing;
    (d) \badaptshort{}+RE constraint with $\kappa=0.08$;
    (e) \badaptshort{}+RE constraint with $\kappa=0.06$;
    (f) \badaptshort{}+RE constraint with $\kappa=0.1$;
    (g) \badaptshort{}+FRE constraint with $\kappa=0.0625$.
    }\label{fig.samples.draw.celeba}
\end{figure}

\begin{figure}
    \centering
    \begin{subfigure}[t]{0.1\textwidth}
        \centering\includegraphics[width=1.2cm]{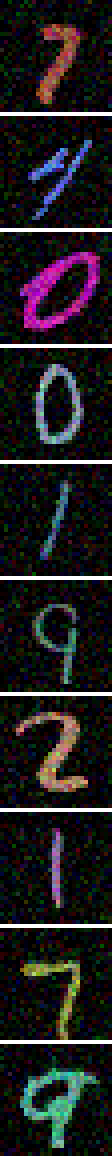}
        \caption{}
    \end{subfigure}
    \begin{subfigure}[t]{0.1\textwidth}
        \centering\includegraphics[width=1.2cm]{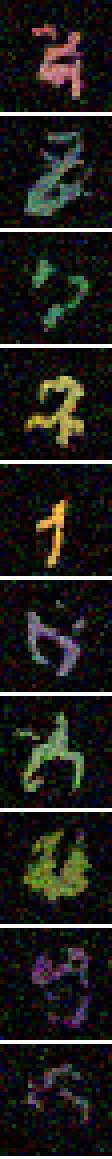}
        \caption{}
    \end{subfigure}
    \begin{subfigure}[t]{0.1\textwidth}
        \centering\includegraphics[width=1.2cm]{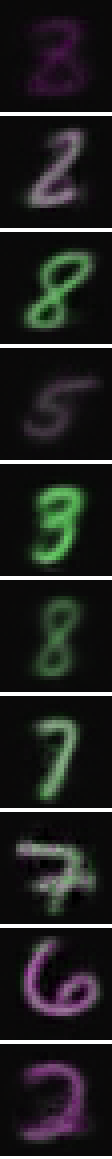}
        \caption{}
    \end{subfigure}
    \begin{subfigure}[t]{0.1\textwidth}
        \centering\includegraphics[width=1.2cm]{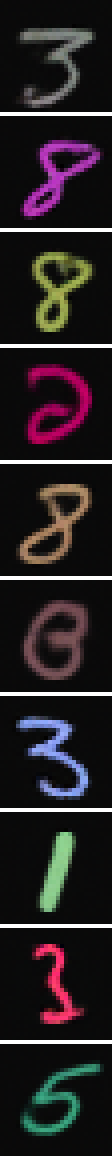}
        \caption{}
    \end{subfigure}
    \begin{subfigure}[t]{0.1\textwidth}
        \centering\includegraphics[width=1.2cm]{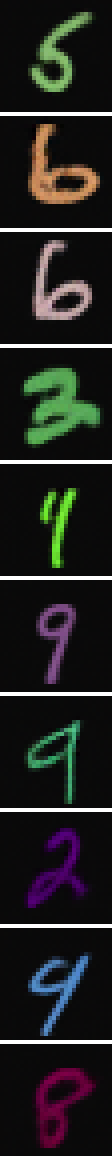}
        \caption{}
    \end{subfigure}
    \begin{subfigure}[t]{0.1\textwidth}
        \centering\includegraphics[width=1.2cm]{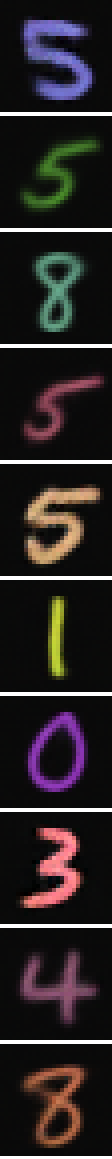}
       \caption{}
    \end{subfigure}

    \caption{
    {\bf Samples from ConvDraw trained on Color-MNIST}. In each block of samples, rows correspond to samples from the data, model reconstructions and model samples respectively.
    From left to right we have models trained with: 
    (a) Data;
    (b) ELBO only; 
    (c) ELBO + Hand crafted $\beta \in [2.0, 0.7]$ annealing;
    (d) \badaptshort{}+RE constraint with $\kappa=0.08$;
    (e) \badaptshort{}+RE constraint with $\kappa=0.06$;
    (f) \badaptshort{}+RE constraint with $\kappa=0.1$.
    }\label{fig.samples.draw.colormnist}
\end{figure}

\begin{figure}
    \centering
    \begin{subfigure}[t]{0.1\textwidth}
        \centering\includegraphics[width=1.2cm]{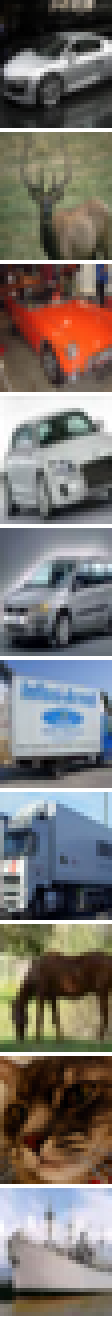}
        \caption{}
    \end{subfigure}
    \begin{subfigure}[t]{0.1\textwidth}
        \centering\includegraphics[width=1.2cm]{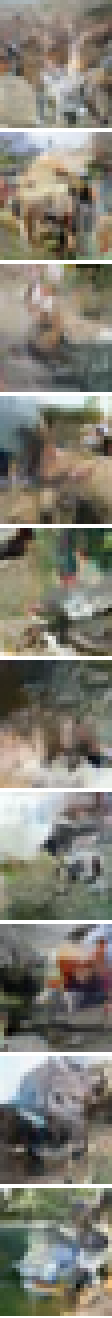}
        \caption{}
    \end{subfigure}
    \begin{subfigure}[t]{0.1\textwidth}
        \centering\includegraphics[width=1.2cm]{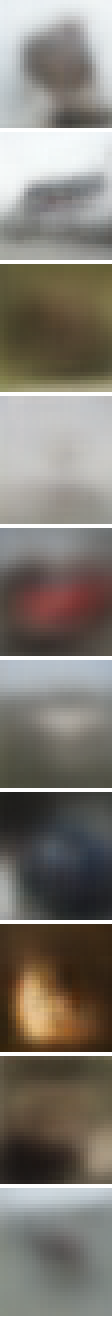}
        \caption{}
    \end{subfigure}
    \begin{subfigure}[t]{0.1\textwidth}
        \centering\includegraphics[width=1.2cm]{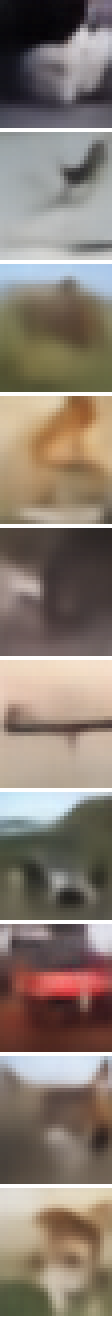}
        \caption{}
    \end{subfigure}
    \begin{subfigure}[t]{0.1\textwidth}
        \centering\includegraphics[width=1.2cm]{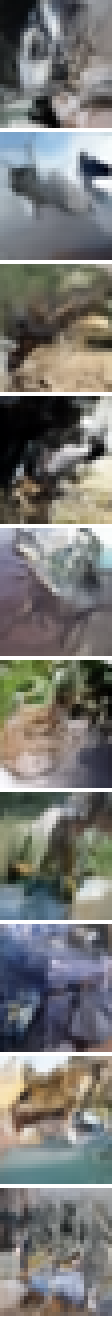}
       \caption{}
    \end{subfigure}
    \begin{subfigure}[t]{0.1\textwidth}
        \centering\includegraphics[width=1.2cm]{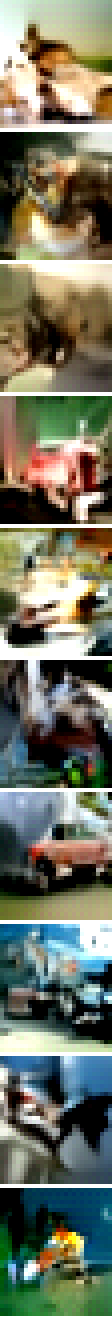}
       \caption{}
    \end{subfigure}
    \begin{subfigure}[t]{0.1\textwidth}
        \centering\includegraphics[width=1.2cm]{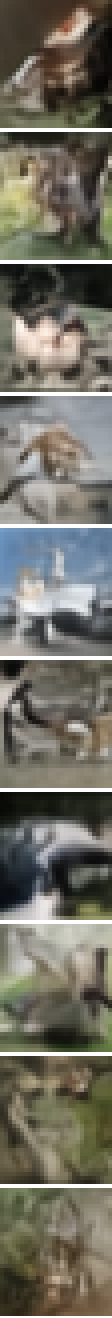}
       \caption{}
    \end{subfigure}

    \caption{
    {\bf Samples from ConvDraw trained on CIFAR10}. In each block of samples, rows correspond to samples from the data, model reconstructions and model samples respectively.
    From left to right we have models trained with: 
    (a) Data;
    (b) ELBO only; 
    (c) ELBO + Hand crafted $\beta \in [2.0, 0.7]$ annealing;
    (d) \badaptshort{}+RE constraint with $\kappa=0.06$;
    (e) \badaptshort{}+RE constraint with $\kappa=0.0028$;
    (f) \badaptshort{}+FRE constraint with $\kappa=0.0625$;
    (g) \badaptshort{}+pNCC constraint.
    }\label{fig.samples.draw.cifar10}
\end{figure}